# An End-to-End Learning Approach for Trajectory Prediction in Pedestrian Zones

Ha Q. Ngo, Christoph Henke, Frank Hees
Institute for Management Cybernetics, Aachen, Germany
*{quangha.ngo, christoph.henke, frank.hees}@ifu.rwth-aachen.de*

## ABSTRACT

*This paper aims to explore the problem of trajectory prediction in heterogeneous pedestrian zones, where social dynamics representation is a big challenge. Proposed is an end-to-end learning framework for prediction accuracy improvement based on an attention mechanism to learn social interaction from multi-factor inputs.*

## 1. INTRODUCTION

While walking in pedestrian zones such as shopping centers, train stations or sidewalks, people have the ability to obey a large number of common sense rules to avoid obstacles. At the same time, people interact with others in the neighborhood to perform safe and legible trajectories [1, 2]. The situation becomes more complex in heterogeneous scenario, where the space is shared among different moving agent categories (i.e. pedestrians, cyclists, skaters, etc.) with different shapes, dynamic constraints, and behaviors [3].

For autonomous robot navigation in such environments, social interactions modeling is crucial to predict future trajectories of surrounding agents. Figure 1 illustrates social interaction in a human-robot crossing space. Since crowded public areas are usually dense and highly dynamic, automatic trajectory prediction is an extremely challenging problem, which has attracted considerable numbers of research works over the past few years [4].

Traditional methods such as *Social Forces* [5] use hand-crafted functions to model human-human interaction, which fails in real-word social environments. Other methods using *Markov Decision Processes* aim to model human motion characteristics from real-world trajectories and predict future trajectories [6, 7]. However, these methods target smooth motion paths, while human motions are multimodal and socially compliant. Recently, learning-based methods have been applied to train models from real-world data. Imitation learning can effectively model human behaviors using *Inverse Reinforcement Learning* [8-11]. The disadvantage is this method does not consider social interaction, which strongly affects the human trajectory in crowds. In order to solve this problem, several works based on *Recurrent Neural Networks* (RNN) have demonstrated the ability to learn social interactions for more accurate trajectory prediction by capturing human-human relationships in the neighborhood [2, 3, 12-17]. These interactions, in reality, dependent on several visual factors (e.g. human position, velocity, time-to-collision, acceleration, head pose, etc.) [14] and acoustic factors (e.g. surrounding sounds intensity) [18]. However, most of the attention-based learning methods attempt to model social interaction based on a single visual factor (e.g., human position) [4].

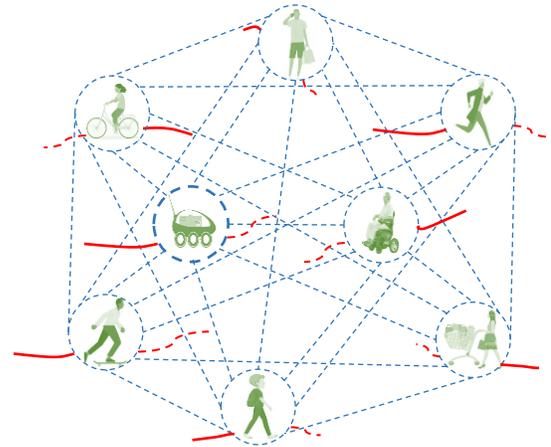

**Figure 1:** Illustration of trajectory prediction in pedestrian zones: green lines - pairwise interactions; red solid lines – past trajectories, red dashed lines – trajectories predicted by the robot.

This paper presents an end-to-end trajectory prediction learning framework architecture for autonomous robot applications in pedestrian zones. We show the ability of our architecture to handle multiple interactive factors from both visual and acoustic stimuli.

## 2. TERMINOLOGY AND BACKGROUND

An end-to-end learning framework for human trajectory prediction, based on the current literature, passes through several steps such as feature extraction, feature encoding, social attention, and trajectory prediction.

*Feature extraction* is the process to capture implicit and/or explicit features of each agent and its surrounding environment from input data. In order to learn the implicit spatial features in the scene, a model based on *Convolutional Neural Network* (CNN) can be applied. The explicit features can also be fed from labeled inputs or by performing object detection and tracking [19-23].

*Feature encoding* process aims to represent the feature dependencies of data. Since human trajectory can be considered as time sequence data, RNN-based architectures such as *Long-Short Term Memory Networks* (LSTMs) can be used to encode human trajectory. LSTM is able to learn dependencies and generalize the properties of temporal sequence data (e.g. past trajectories) [24]. In fact, LSTM-based methods have proved its successes in the area of trajectory prediction in recent years [2, 3, 13, 15, 25].

*Attention mechanism* is the way to learn the importance of the factors in the environment i.e. to capture the interaction between agents. In this context, modeling of social dynamics in crowded scenes is a challenging task as the behaviors of each agent are affected by surrounding agents. Since vanilla LSTM was designed to encode features of an isolated sequence, it fails to learn interaction between agents. Therefore, it is required to apply an *Attention Mechanism* [26], which can share and combine information through different encoders to build a compact representation of the social context. In order to do this, the hidden states from LSTM encoders can be connected to each other through a *Social Pooling* layer [2]. The social interaction can also be represented using LSTM-ConvNet architecture [3]. In this method, the hidden states from LSTM encoders are pushed though convolutional layers to build an implicit compact representation vector. Another alternative is Graph-based method using *Structured-RNN*, which has also demonstrated significant results [14]. This method extends *Graph Attention Network* (GAT) to learn the temporal-space edge factors between the agents [17].

*Trajectory prediction* is the last step to output the future predicted trajectories. In order to maps the representation vector back to target sequences (i.e. predicted trajectories) a *Decoder LSTM* can be used [3, 17, 25, 27, 28]. However, this method is data consuming and leads to the generalization problem. As a solution, *Generative Adversarial Networks (*GANs) [29] can be applied to maximize the lower bound of training data likelihood. This method has shown promising results in several applications including trajectory prediction [13, 25].

## 3. PROPOSED METHOD

In this work, we focus on the attention-based methods for end-to-end human trajectory prediction in heterogeneous scenarios. We consider both visual and acoustic inputs for a comprehensive representation of social dynamic environments. In order to improve prediction accuracy, we propose a novel learning framework architecture, which consists of the following main modules: 1 – Feature extraction; 2 – Feature encoding; 3 – Attention mechanism; 4 –Trajectory prediction (Figure 2).

### 3.1. Feature extraction

We investigate research on how real-world environment stimuli affect human trajectory. Since we attempt to make the robot have the ability to understand the social environment as a human, we consider both visual and acoustic stimuli. Intuitively, when walking in crowds, people use their eyes to perceive visual stimuli (i.e. surrounding obstacles and moving agents) their ears for surrounding sounds mapping. In most cases, to perform safe and legible trajectories, people do not try to capture all the detail information about the surrounding environment, but pay attention to some important factors (distance from neighbor agents, their speed, size, head poses, alarm sounds, etc.) [14]. Therefore, the first module of our framework architecture consists of three different feature extractors for essential input feeding: 1) The Dynamic extractor captures dynamic features from raw visual input using *Tracking by Detection* method with YOLO [20] and Reciprocal Velocity Obstacles (RVO) models [30]; 2) The scene extractor captures the implicit features from the visual scene using VGG network [31]; 3) The Acoustic mapping module also use a CNN model to capture the implicit features regarding the sound categories, intensity, and location of origin.

### 3.2. Feature encoding

Since human trajectory can be considered as a temporal sequence, it can be encoded using a multi-layer LSTM module [24]. Each encoder LSTM captures the motion features of one agent in the relationships with its visual stimuli. This way, the temporal features of each agent are represented as a hidden state vector [2].

### 3.3. Attention mechanism

In dense space, human trajectory is frequently affected by other agents in the neighborhood. However, the hidden state vector of an encoder LSTM is a temporal representation of only one agent. It does not capture the interaction with other agents. Therefore, in this work, we use a *Social Pooling* layer to learn the interaction between agents [2]. The social Pooling layer is applied for all three attention learning modules (spatial, dynamic and acoustic attention) to capture interactive relationships between agents in the form of hidden state vectors. These vectors are later on concatenated into a compact vector, which is the implicit representation of social interaction.

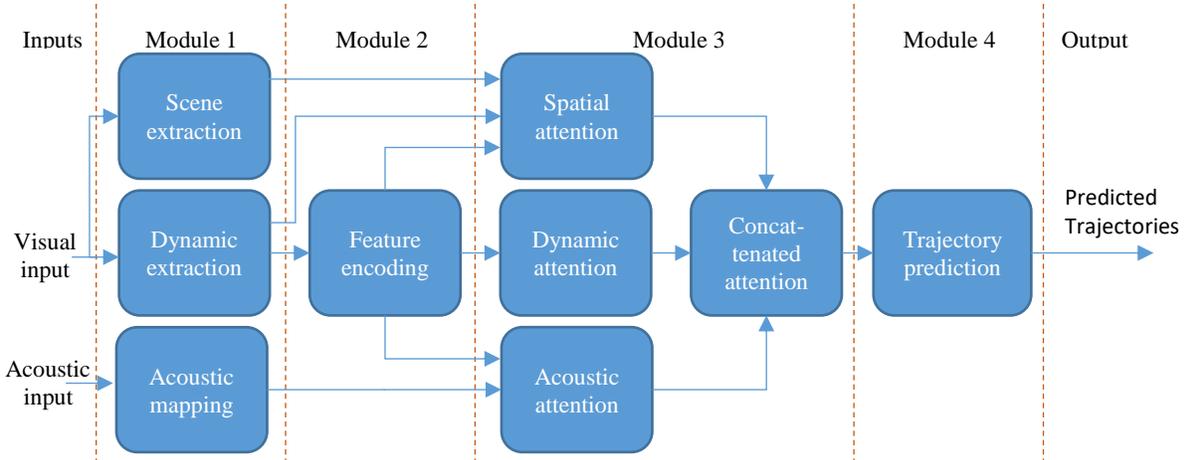

**Figure 2:** Proposed framework architecture: Module 1 – Features extraction; Module 2 – Features encoding; Module 3 – Social interaction learning; Module 4 – Future trajectory prediction

### 3.4. Trajectory prediction

At the final step of the framework, an LSTM-based GAN module [13] is applied to predict the future trajectory for each agent. GAN module consists of two components: 1) The G*enerator* learns the distribution of the trajectories and generates a sample of the plausible future trajectories for each agent using a decoder LSTM; 2) The discriminator learns to distinguish whether a generated trajectory is feasible or infeasible using a classifier LSTM. This min-max game is trained simultaneously until the distinguishing ratio is 50/50.

## 4. DISCUSSION AND FUTURE WORK

Introducing acoustic input data into the learning process makes the opportunity for a better understanding of the surrounding environment. However, this can raise the cost for real-time applications since a sound localization system is required. In addition, the learning framework requests synchronous training data of visual and acoustic inputs. Unfortunately, such a dataset for public crowded space scenario is still not available. In the next step, we propose to create a new dataset for this research domain.

Another problem is the complexity of pedestrian zones. Instant trajectory prediction of multiple interactive agents is a computational time consuming task. To overcome this problem, we consider reducing the number of agents using a grouping model [32]. This way, social interaction can be considered as the interaction between groups. However, since people frequently switch between groups, we consider solving the problem of features encoding at grouping/ungrouping moments.

## 5. CONCLUSSION

We presented an end-to-end learning framework architecture for trajectory prediction in pedestrian zones. The model architecture is able to consider both visual and acoustic inputs. The advantages of CNNs, LSTMs, and GANs were inherited in our architecture for the environment feature extraction, feature encoding, and trajectory prediction steps respectively. The attention mechanism is able to learn from different kinds of factors in the social dynamic environment and to capture the interaction between agents, which can improve the prediction accuracy. Related issues and proposed solutions were also discussed.

### ACKNOWLEDGMENT

This work was conducted as part of the interdisciplinary project UrbANT, which is supported from the German Federal Ministry of Education and Research under the funding code 16SV7919.